\documentclass[letterpaper, 10 pt, conference]{ieeeconf}  

\IEEEoverridecommandlockouts                              

\usepackage{amsmath}
\usepackage{graphicx}
\usepackage{wrapfig}
\usepackage{caption}
\usepackage[hidelinks]{hyperref}
\usepackage{cleveref}
\usepackage{booktabs}
\usepackage{xcolor}
\usepackage[sort, compress]{cite}

\title{\LARGE \bf
CORL-AUV: CFD Oriented Reinforcement Learning\\%
for Autonomous Underwater Vehicles
}

\author{Steven Roche$^{1,\ast}$, Milo Van Mooy$^{2}$, Nathan McGuire$^{2}$, Levi Cai$^{1}$, Jonathan P. How$^{3}$, Yogesh Girdhar$^{2}$   
\thanks{*This work was supported by the DoD SMART Scholarship.}
\thanks{$^{1}$S.\ Roche and L.\ Cai are with MIT-WHOI Joint Program, MA 02543, USA (e-mail: \{rochesh,cail\}@mit.edu).}%
\thanks{$^{2}$M.\ Van Mooy, N.\ McGuire, and Y.\ Girdhar are with WHOI, MA 02543, USA (e-mail: milovm@stanford.edu, \{nmcguire,ygirdhar\}@whoi.edu).}%
\thanks{$^{3}$J.\ How is with Massachusetts Institute of Technology, Cambridge, MA 02319, USA (e-mail: jhow@mit.edu).}%
}

\begin{document}
\maketitle
\thispagestyle{empty}
\pagestyle{empty}

\begin{abstract}
Precise control and positioning of autonomous underwater vehicles (AUVs) is critical for sampling, maintenance, and survey applications. Reinforcement learning (RL) offers rapid controller development while handling a range of deployment parameters via domain randomization (DR). However, DR is limited by the underlying simulation's capacity to model real physics such as drag, which is a large contributor to sim-to-real gaps. Computational fluid dynamics (CFD) provides high-fidelity drag models but has significant computational cost. Thus, in this paper we train surrogate approximations of CFD data of a given vehicle, providing drag estimates $700,000\times$ faster than solving for drag at each time step with CFD within the RL training pipeline. We deploy a Proximal Policy Optimization (PPO) policy zero-shot on a 6-DOF AUV in which policy training is performed on these surrogate drag models (SDMs), providing analysis on zero-shot transfer, reward shaping sensitivity, and physical parameter perturbations. On 15 identical segments in the ocean, we observe median gains of $19\%$ in mean tracking error and $31\%$ in control effort compared to the equivalent inertia box model. Our SDM based RL controller better predicts zero-shot transfer and reaches the most waypoints across all three reward shaping choices that we evaluate. Finally, we add $0.907$ $kg$ to the vehicle perturbing the mass, center-of-mass, and center-of-buoyancy. The policy reaches $100\%$ of waypoints only when the SDM drag model is used; the policy fails to reach any of the 15 waypoints when the other drag models are used. 
\end{abstract}

\section{INTRODUCTION}
Autonomous underwater vehicles (AUVs) have been used for a variety of underwater environmental research interests, from the study of plankton \cite{c1}, tracking of mobile animal species \cite{c2,c3}, and surveys of coral reefs \cite{c4}. These tasks often require robots to navigate challenging terrain and conditions  \cite{c5,c6}; for example, a vehicle may need to operate within sub-meter ranges of a fragile coral species to obtain high-resolution visual data \cite{c4, c7}, or it may need to use multiple sensors to avoid obstacles \cite{c8}. Maintaining such close proximity to obstacles and reacting quickly to marine life requires fine-grained control of the dynamics of the vehicles. Complicating matters, a single at-sea mission may require multiple sensor configurations making controller development extremely time-intensive and challenging due to the changes in the hydrodynamics. Approaching this problem with traditional controllers such as PID often leads to controllers that need re-tuning, fail to generalize to new tasks, or degrade under changing environmental conditions. Recent works have turned to reinforcement learning (RL) based controllers to address these limitations and have shown superior performance over PID \cite{c11_anonymized, swim4Real}. Two common approaches to dynamics modeling involve: 1) using high-fidelity computational fluid dynamics (CFD) solvers directly in the RL training loop to obtain an accurate controller \cite{pointToPointNavigationFishSwimmer, RLforTurbulentFlows, Cui_2024, learningFrameworkForFishRobots, lin2025learningagileswimmingendtoend}, or 2) using analytical low-fidelity models to obtain a reasonable controller within minutes \cite{c11_anonymized, swim4Real, tunçay2026fastpolicylearning6dof, fosso2025sim2swimzeroshotvelocitycontrol,weng_autonomous_2024,liu_deep_2021}. For a 6 degree-of-freedom (6-DOF) vehicle, exhaustive CFD sampling quickly becomes computationally prohibitive. If each component of the linear- and angular-velocity is sampled $n$ times over an expected operating range, the resulting CFD dataset scales as $O(n^6)$. Each computation could take dozens of seconds or even minutes to solve a transient flow case. While this exhaustive CFD approach produces an accurate controller in a zero-shot manner, it is challenging to rapidly adapt to new vehicle parameters or update the controller design without expensive re-training. In this work, we effectively distill complex CFD models with neural network approximations, achieving fast inference while reducing the number of CFD samples required to train the policy. We refer to these approximations as surrogate drag models (SDMs), which can be incorporated into RL frameworks to rapidly create new controllers while benefiting from CFD. 

\begin{figure*}[t]
    \centering
    \includegraphics[width=1\linewidth]{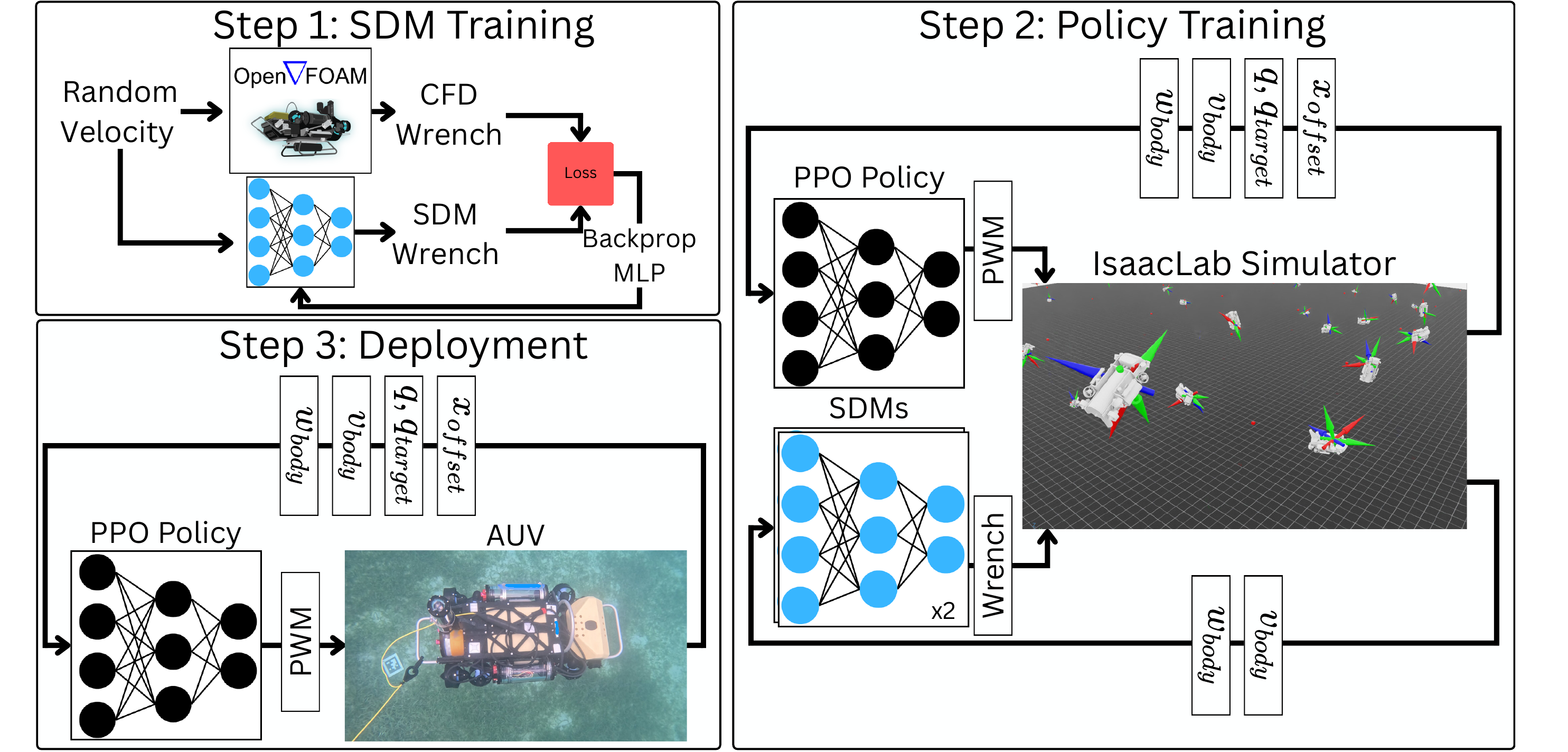}
    \caption{Overview of our method that incorporates surrogate models directly into IsaacSim. Upper left: SDM training approach. Bottom left: policy controlling vehicle in a reef environment. Right side: RL policy training approach.}
    \label{fig:overviewMethod}
\end{figure*}

Domain randomization alone may fail to compensate for important inaccuracies in a nominal dynamics model, as evidenced by our results from the commonly used equivalent inertia box drag model and a system identification (ID) drag modeling approach. 
We show that increasing model fidelity through computationally efficient CFD surrogates improves not only zero-shot performance, but also the reliability with which simulation predicts successful transfer to hardware. Our contributions are as follows:
\begin{enumerate}
\item A CFD-surrogate approach enabling zero-shot sim-to-real RL for a 6-DOF AUV, with median improvements over the equivalent inertia box baseline of $19$\% in mean tracking error, $11$\% in traversal time, and $31$\% in control effort.
\item Evidence that increased drag-model fidelity reduces sensitivity to reward shaping and cannot necessarily be replaced by domain randomization. Across three sets of reward-shaping coefficients, the SDM-based policy reaches more waypoints than policies trained with the lower-fidelity drag models.
\item Demonstration that higher model fidelity improves the predictive value of simulation for sim-to-real transfer. Under an additional $0.907$ kg payload, simulation correctly predicts successful transfer of the SDM-based policy, which reaches $100$\% of hardware waypoints, while the simulator incorrectly predicts that the system ID-based policy would transfer.
\item Experimental demonstration of zero-shot transfer beyond controlled tank experiments to a real-world underwater environment. We train the SDM-based PPO policy in simulation to execute visual surveying patterns and deploy it successfully in Lameshur Bay, St. John, USVI.
\end{enumerate}

\section{RELATED WORKS}
\subsection{Reinforcement Learning for AUV Control}
\label{sec:rlForAUVControl}
While various control paradigms exist for AUVs, such as PID, model predictive control \cite{c9}, lead-lag \cite{c10}, and sliding mode \cite{c10}, recent work has turned to reinforcement learning to avoid manual tuning often required for a new vehicle configuration. Many of these \cite{c11_anonymized, swim4Real, tunçay2026fastpolicylearning6dof, fosso2025sim2swimzeroshotvelocitycontrol} use MuJoCo's equivalent inertia box \cite{Todorov2012MuJoCoAP}, a simplified hydrodynamics model that may prevent controllers from learning higher performance or more energy efficient maneuvers. On a 6-DOF AUV, Anonymous \cite{c11_anonymized} deployed a zero-shot RL policy that successfully station keeps while recovering from disturbances. Swim4Real \cite{swim4Real} deployed a position controller on a 6-DOF AUV, performing extensive in-water position and orientation experiments and demonstrating a $39\%$ reduction in energy usage compared to PID. Tuncay et al. \cite{tunçay2026fastpolicylearning6dof} evaluated three RL algorithms on a 6-DOF vehicle and compared to an MPC baseline, and Fosso et al. \cite{fosso2025sim2swimzeroshotvelocitycontrol} trained a BlueROV2 Heavy within 3 minutes and corrected steady-state errors in the controller with an integral term.

\subsection{CFD for Learning-based Control}
\label{sec:CFDwithRLforAUVControl}
Besides domain randomization, a strategy to close the sim-to-real gap is to make a model of the platform and simulate the hydrodynamics with CFD. Zhu et al. \cite{pointToPointNavigationFishSwimmer} used the lattice Boltzmann method with a deep recurrent Q-network to simulate a fish robot reaching goal poses in a flow field. Lidtke et al. \cite{RLforTurbulentFlows} used Partially Averaged Navier Stokes with KSKL to train an RL controller on a 2D version of the BlueROV2 Heavy, which enabled vehicle control in simulations of various flow conditions. Cui et al. \cite{Cui_2024} learn a deep RL controller with CFD in an end-to-end manner to control a simulated 4-DOF robotic fish, discovering three distinct motion patterns. Zhang et al. \cite{learningFrameworkForFishRobots} first train an RL controller for a 3-DOF fish robot in an environment built using experimental data and then refine the policy in a CFD environment. They deploy the resulting policy on a real fish robot for a path following and pose attainment task. Lin et al. \cite{lin2025learningagileswimmingendtoend} develop an end-to-end deep RL pipeline utilizing the FishGym CFD simulator \cite{liu2022fishgymhighperformancephysicsbasedsimulation} which enables the policy to directly output actuator commands on a real 3-DOF fish robot. Sun et al. \cite{SUN2026125161} learn a CFD-derived transition model of a 1-DOF fish whose control input is oscillation frequency; they learn varied swim strategies using PPO. Recent work by Xu et al. \cite{11509187} learns their vehicle's 6-DOF hydrodynamic forces and moments (the wrench), which is used in a rigid-body hydrodynamics model. They demonstrate smooth trajectory following in sim with a non-singular terminal sliding-mode controller. 

We likewise learn CFD-derived surrogate models of the hydrodynamic wrench, but use them within a highly parallel RL environment to train a PPO policy that directly outputs thruster commands from pose-error and body-frame velocity observations. For low-speed operation, we reduce the CFD sampling burden by decoupling translational and rotational velocity sampling while retaining cross-axis coupling within each subspace. We then evaluate the resulting policies on hardware, probing zero-shot sim-to-real transfer, sensitivity to reward shaping, and robustness to physical parameter perturbations.

\section{METHOD}
We fit SDMs on steady-state CFD data to enable fast approximation of the results within the policy training loop. As outlined in ~\ref{sec:CFDSubsection}, we independently solve linear- and angular-velocity cases, reducing the number of samples needed to cover the vehicle's operating range at the cost of approximation error. We modify Anonymous' \cite{c11_anonymized} training pipeline to incorporate these new models (Figure~\ref{fig:overviewMethod}). We investigate how the following drag model assumptions affect sim-to-real transfer: 
\begin{enumerate}
    \item The equivalent inertia box drag model, as a baseline comparison (diagonal drag model)
    \item System ID on real vehicle data (diagonal drag model)
    \item Surrogate drag models that approximate the steady state CFD data (coupled axis model)
\end{enumerate} 
Including the equivalent inertia box and a system ID approach enables comparison between two common methods while fixing their assumption on the drag structure-a diagonal matrix. The SDM and equivalent inertia box methods both require no hardware testing to obtain, providing a comparison of simulation methods. Comparing SDM to system ID provides some insight into how modeling cross-coupling may affect sim-to-real transfer, although system ID having access to real vehicle data limits this comparison.

\subsection{Drag Modeling and Dynamics Assumptions}
Domain randomization with parallelization is a widespread method of learning robust controllers in simulation and dealing with the sim-to-real gap \cite{reinforcementLearningSurvey2026}. Using RL with a simulator can learn coupled and robust controllers while achieving energy benefits \cite{swim4Real}. While the works in section~\ref{sec:rlForAUVControl} handle variation in the vehicle's mass, center-of-mass (COM) and center-of-buoyancy (COB) via DR, they do not quantify nor handle the uncertainty in their drag modeling. Hence, we explore how these drag models and our proposed model affect sim-to-real transfer. We provide more details of the models below:

\subsubsection{Equivalent inertia box}
\label{sec:inertiaBox}
The equivalent inertia box method estimates the forces and torques given the mass and inertia of the vehicle by assuming the vehicle maintains a box-shape with equivalent inertial characteristics. For compactness, we do not list the full equations here, but see MuJoCo \cite{Todorov2012MuJoCoAP} and Anonymous \cite{c11_anonymized} for details.

\subsubsection{System ID}
\label{sec:systemID}
\begin{figure}[t]
  \centering
  \includegraphics[width=\linewidth]{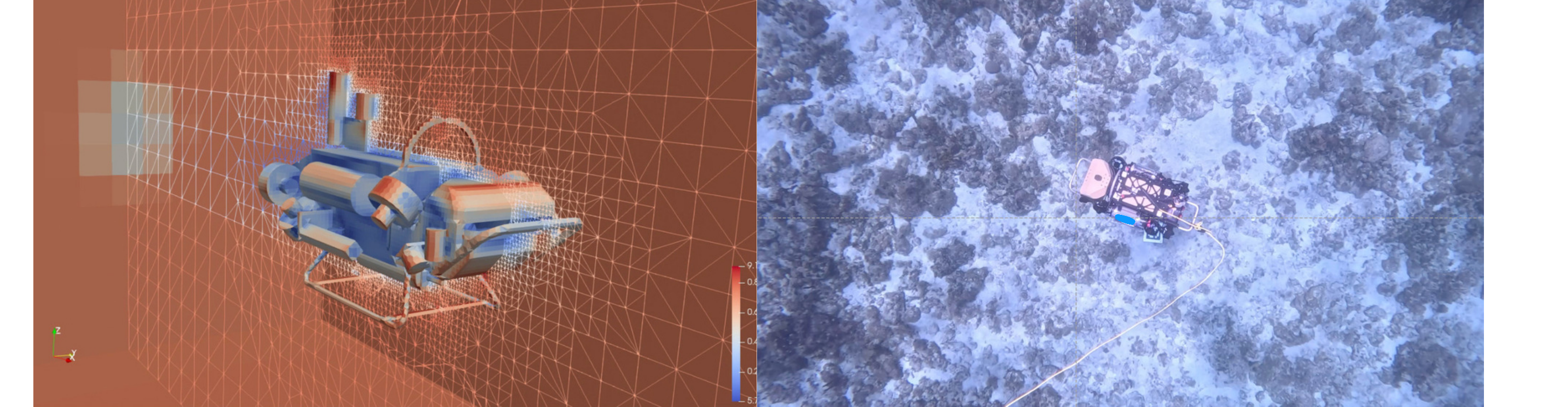}
  \caption{Left: OpenFOAM solving the fluid velocity on the vehicle. Right: Field deployment at Yawzi Reef, USVI.}
  \label{fig:cfdExample}
\end{figure}
For system ID, we assume a linear diagonal drag model, in which each dominant translational or rotational axis is modeled independently, neglecting cross-axis coupling. We neglect cross-axis coupling because of the significant challenge of repeatedly inducing coupled motions on the real vehicle. We estimate the drag coefficients from coast-down data collected on the real vehicle using IMU and DVL measurements fused with an Extended Kalman Filter. A coast-down trial consists of driving the vehicle to an initial velocity along a dominant axis of motion, cutting the thrusters, and allowing the vehicle to drift freely toward zero velocity. For each translational axis \(i\) and for each rotational axis \(j\) we fit:
\begin{align}
m_{\mathrm{eff},i}\frac{dv_i}{dt} = -d_i v_i,\quad I_j\frac{d\omega_j}{dt} = -k_j \omega_j.
\end{align}
where $m_{\mathrm{eff},i}$ is the effective mass along the $i$-th component, $I_j$ is the moment of inertia about the $j$-th axis, and $v_i$ and $\omega_j$ represent the linear- and angular-velocities. We estimate and remove the vertical-velocity bias induced by buoyancy. The ODEs are simulated with fourth-order Runge--Kutta integration and coefficients are optimized to minimize the error between predicted and observed velocities. Coefficients are fit independently using data along the corresponding \(\pm i\) translational axis or clockwise/counterclockwise rotational axis. We collect 60 translational trials, 10 along each of \(\pm x\), \(\pm y\), and \(\pm z\), and 30 rotational trials, 5 clockwise and 5 counterclockwise for roll, pitch, and yaw; one yaw trial was corrupted, so 29 rotational trials were used for fitting. This data collection approach unfortunately biases system ID to fit slower speeds. We generated a steady-state CFD dataset that solves cases involving both linear and angular velocities (fully coupled in 6-D space) and found that this method predicts drag better than the equivalent inertia box until speeds increase beyond about $0.43$ $m/s$.

\subsubsection{Computational Fluid Dynamics}
\label{sec:CFDSubsection}
\begin{table*}[t]
  \centering
  \caption{Drag-model RMSE and mean absolute error relative to CFD data where we solve both linear- and angular-velocities in the same case.}
  \label{tab:drag-model-errors}
  \small
  \setlength{\tabcolsep}{4pt}
  \begin{tabular}{l c c c c c}
    \toprule
    \textbf{Model} & \textbf{Load} & \multicolumn{2}{c}{\textbf{All observations} ($n=9{,}125$)} & \multicolumn{2}{c}{\textbf{Lower 75\% velocity range} ($n=5{,}124$)} \\
    \cmidrule(lr){3-4}\cmidrule(lr){5-6}
    & & \textbf{RMSE} & \textbf{Mean error} & \textbf{RMSE} & \textbf{Mean error} \\
    \midrule
    SDM & Force ($N$) & 15.03 & 8.28 & 8.36 & 4.62 \\
     & Moment ($N\,m$) & 1.794 & 0.899 & 0.981 & 0.497 \\
    \addlinespace
    Equiv.\ Inertia Box & Force ($N$) & 29.33 & 18.47 & 16.46 & 10.35 \\
     & Moment ($N\,m$) & 3.184 & 2.167 & 1.772 & 1.213 \\
    \addlinespace
    System ID & Force ($N$) & 41.22 & 27.19 & 21.30 & 13.88 \\
     & Moment ($N\,m$) & 3.173 & 2.145 & 1.737 & 1.165 \\
    \bottomrule
  \end{tabular}
\end{table*}
In contrast with the other baseline drag models, the CFD data captures cross axis coupling in its simulations. But sampling over the entire operating range can be computationally prohibitive for the 6-DOF setting; by solving the hydrodynamic effects of linear- and angular-velocities separately, we reduce the sampling from 6-D space to 3-D space. A sample takes about $100$ $s$ in 3-D space, requiring $2$ days to produce $10^4$ samples for each dataset with $48$ CPUs. An equivalent coverage in 6-D space requires $10^4 \times 10^4 = 10^8$ samples, which would take decades to compute.
Thus, using OpenFoam \cite{noauthor_openfoam_2025} with our vehicle's CAD model (Figure~\ref{fig:cfdExample}), we randomly sample a linear- or angular-velocity for each case and solve for the steady-state wrench with the PIMPLE algorithm. We use RANS K-$\omega$ SST \cite{tensorialApproachCompCont} as our turbulence model. This produces two datasets where the features are a linear-velocity with an associated wrench or an angular-velocity with an associated wrench. We obtain $\sim$183,000 examples for linear velocities and $\sim$90,000 examples for angular velocities. We fit an MLP on each dataset. During inference, the outputs of the MLPs are summed together to produce the estimated wrench at each time step. 

While this method retains the cross-axis coupling within each linear- or angular-velocity case, this decomposition does not model some coupling effects. The RMSE and mean absolute error for each drag model in comparison to a CFD dataset solving linear- and angular-velocities together in the same case can be seen in Table~\ref{tab:drag-model-errors}.
\begin{figure*}[t]
    \centering
    \includegraphics[width=1\linewidth]{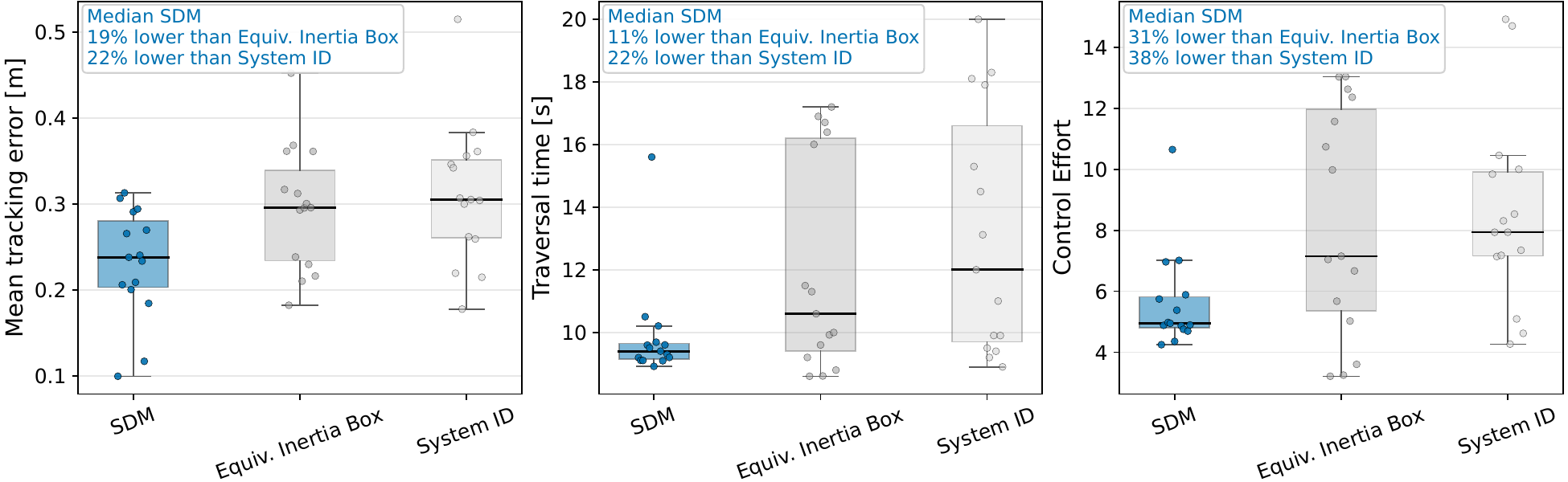}
    \caption{On a U-Shape pattern conducted in the field, we find that the SDM-based RL policy results in lower mean tracking error, traversal completion times, and control effort. Each U-shape contains 3 segments of identical length and we run the experiment 5 times for each drag model (15 segments per model, the results of which are shown here).}
    \label{fig:fieldUResults}
\end{figure*}
\subsection{Domain Randomization}
\label{sec:domainRandomization}
As in prior works (\ref{sec:rlForAUVControl}), we seek policies that are robust to small parameter perturbations to the COM and COB which often result from slight changes in sensor configurations. Following \cite{molchanov2019simtomultirealtransferlowlevelrobust, c11_anonymized}, we utilize domain randomization (DR) on the volume and COB-COM offset of the vehicle. For the ocean deployments, we use a small DR range of $\pm0.01$ $m$ for the COB-COM offset and $\pm 1\%$ of the volume of the vehicle. With these DR parameters, a vector sampled uniformly from a sphere of radius $0.01$ $m$ is added to the default COB-COM offset and the volume is set within $1\%$ of $0.025977$ $m^3$. Motivated by the problem of handling new vehicle configurations, we calculate the shift in these parameters when placing weights on the stern of the vehicle. These calculations revealed that to handle $\pm0.453$ $kg$ in mass we should use $\pm0.01$ $m$ for COB-COM offset DR and $\pm1\%$ volume DR. For handling $0.907$ $kg$ in mass we double the DR values.
\subsection{Training Details}
\subsubsection{PPO Training}
Each policy is trained in IsaacSim \cite{NVIDIA_Isaac_Sim} with $10,000$ environments learning in parallel. Each vehicle spawns with a random pose within $5$ $m$ of a goal position. The vehicle is tasked with achieving both the goal position and the goal orientation in $10$ seconds. We train the policy with on-policy implementation (PPO) \cite{schulman_proximal_2017} using the RSL-RL implementation \cite{pmlr-v164-rudin22a}. We perform 5 learning epochs per update with 4 mini-batches. Adaptive KL controls the learning rate schedule and we use an initial learning rate of $5\cdot10^{-4}$. We set $\gamma = 0.99$. At each time step, the vehicle receives an observation of its target orientation, current orientation, body-frame linear-velocity, body-frame angular-velocity, and offset from the goal position. Table~\ref{tab:sim_params} displays the default parameters. The domain randomization used in this work is centered on the Volume and COB-COM offset values listed there. 
\begin{table}[t]
\centering
\caption{Default simulation parameters.}
\label{tab:sim_params}
\begin{tabular}{l c c c}
\toprule
\textbf{Parameter} & \textbf{Symbol} & \textbf{Value} & \textbf{Units} \\
\midrule
Mass & $m$ & $25.90$ & $kg$ \\
Fluid density & $\rho$ & $997.00$ & $kg/m^3$ \\
Fluid viscosity & $\beta$ & $0.001306$ & $Pa\cdot s$ \\
Time step & $\Delta t$ & $0.02$ & $s$ \\
Volume & $V$ & $0.025977$ & $m^3$ \\
COB-COM Offset & COB-COM & $[0.00, 0.00, 0.01]$ & $m$ \\
\bottomrule
\end{tabular}
\end{table} 

\subsubsection{SDM Training}
We trained two MLPs: the first was trained on the linear-velocity dataset and the second was trained on the angular-velocity dataset. Thus the input is a linear- or angular-velocity and the output is a predicted wrench. Each MLP has 2 hidden layers with $1024$ units. We used a learning rate of $2\cdot 10^{-3}$ and $2\cdot 10^{-4}$ for the linear and angular MLP network, respectively. We used a batch size of $64$ for the linear model and a batch size of $32$ for the angular model. We use MSE loss, Adam, ReLU, and a $80/10/10$ training split. We trained for $200$ epochs and selected the best parameters based on the validation set to mitigate overfitting. These lightweight surrogates enable an RL controller to be developed in $15$ minutes or less.
\subsection{Reward Shaping}
\label{sec:trainingApproachandRewardShaping}
\subsubsection{Reward Shaping Sensitivity}
\label{sec:generalTask}
\begin{figure*}[h]
    \centering
    \includegraphics[width=1.0\linewidth]{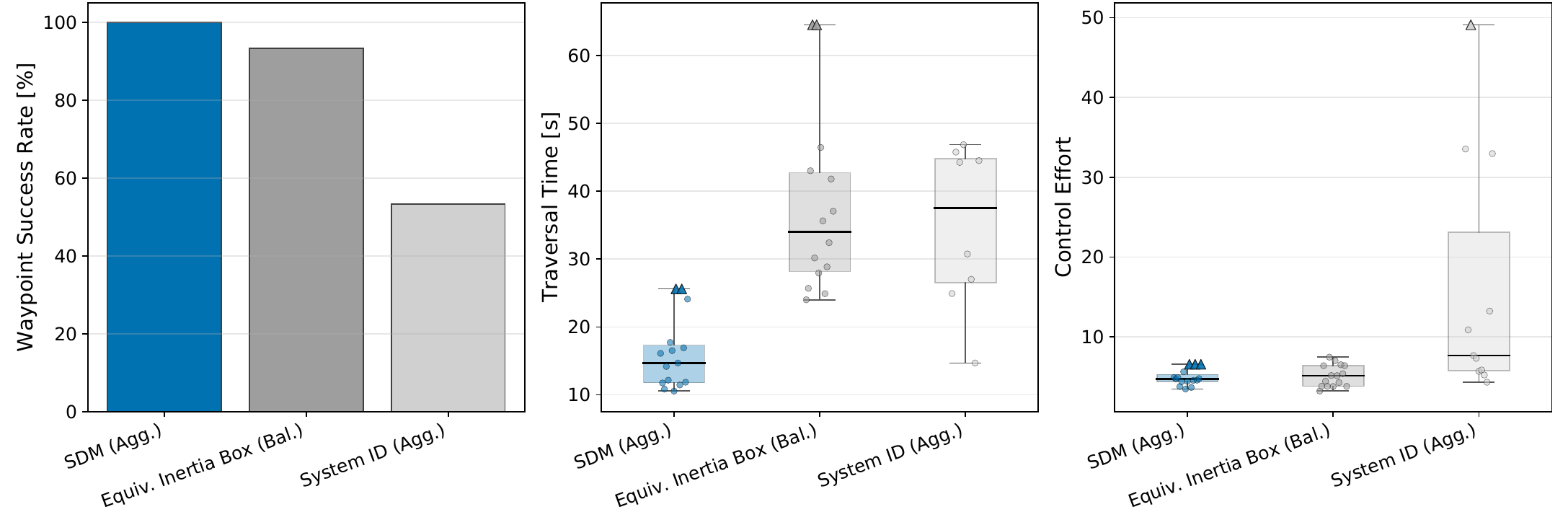}
    \caption{Comparing the best policies discovered during the reward shaping tank experiment, we find the SDM-based RL policy completes the most waypoints, doing so faster and with less energy. Agg. = Aggressive, Bal. = Balanced.}
    \label{fig:tankBestResults}
\end{figure*}
The reward shaping method we test uses Gaussian distributions for each reward term. We intentionally evaluate three reward coefficient sets across all three drag models to assess how sensitive sim-to-real transfer is to reward design, a sensitivity that is often not characterized in RL studies. The vehicle traces the sides of a box with $1.7$ $m$ long sides in a controlled tank environment, achieving a particular 3D pose at each corner before proceeding to the next goal point. The vehicle's initial starting position could bias the results of the first leg, so we only report results on the final $3$ segments which we refer to as the ``U-shaped pattern''. We generate three sets of coefficients, denoted as ``conservative'', ``balanced’’, and ``aggressive’’. No drag model receives model-specific reward tuning or additional hardware feedback. This protocol allows us to evaluate whether the choice of drag model affects robustness to reward shaping under otherwise identical conditions. The reward function is:
\begin{equation}
r_{\mathrm{total}} \;=\; \sum_{m \in \mathcal{M}} \lambda_m e^{-||m||^2},
\qquad \mathcal{M} = \{p,\, q,\, v, \, a\}.
\end{equation}
After this experiment, we select the best performing set for each drag model and train three new policies with DR activated. We use this procedure in an effort to give each model the highest probability of transferring well. We use the $0.907$ $kg$ DR range described in Section~\ref{sec:domainRandomization} to investigate the sim-to-real transfer performance. After training the new policies, we add $0.907$ $kg$ of mass to the stern of the vehicle, and we task the vehicle with tracing the same U-shaped pattern in the tank.
\subsubsection{Zero-Shot RL Visual Survey Deployment}
\label{sec:specificApplication}
We adjust the reward objective function to incentivize behavior that is useful for visual surveys. We task the vehicle with tracing the U-shaped pattern with three $4$ $m$ sides. At the start of all segments the vehicle turns $90^\circ$ clockwise and then travels straight to the goal point. In this particular case, we do not have a goal orientation at the goal point. The objective function is updated to:
  \begin{equation}
  r_{\mathrm{total}} \;=\; \sum_{k \in \mathcal{K}} \lambda_k r_k,
  \qquad \mathcal{K} = \{p,\, q,\, \mathrm{stable},\, v,\, w,\, a\}.
  \label{eq:reward}
  \end{equation}
To encourage fully arriving at the goal point, we define 3 distance-based reward terms, defined as:
$
r_{dist}(b) = \exp\left[-\left(\frac{||p||}{b}\right)^2\right]
$.
We set $b_1 = 2.0$ to give a diffuse position reward signal and add in two terms with $b_2 = 0.35, b_3 = 0.12$ to encourage the model to fully close the distance to the goal.

Because marine visual surveys typically consist of lawnmower patterns that benefit from a stable direction of motion, we further encourage the vehicle to face the goal point while traveling to it. When beyond 1.0 $m$ we gate the position reward with whether the vehicle is facing the target goal, which we define as $r_{\psi} = \exp\left[-\left(\frac{e_\psi}{a}\right)^2\right]$, where $e_\psi$ is the heading error, $a = 0.75$, and $\psi$ represents yaw.
Thus, the total position reward is:
\begin{align*}
r_p = r_{\psi}(r_{dist}(b_1) + \lambda_q) + r_{dist}(b_2) + r_{dist}(b_3).
\end{align*}
The first term contributes the most reward signal at the beginning, encouraging the vehicle to obtain the correct orientation and then drive toward the goal point. When within $1.0$ $m$, the first term drops out. We add an additional stability term to encourage the vehicle to remain upright with minimal variation in pitch and roll. We define:
\begin{align*}
r_{stable} = \exp\left[-\frac{\phi^2 + \theta^2}{0.5^2} \right] + \exp\left[\left(-\frac{1 - \hat{z}_b\cdot\hat{z}_w}{0.35}\right)^2 \right]
\end{align*}
where $\phi$ and $\theta$ are roll and pitch, and $\hat{z}_b\cdot\hat{z}_w$ measures the alignment between the vehicle and world up directions. Finally, we add $r_v$ and $r_w$ to mitigate oscillations, and we add $r_a$ to encourage less thruster usage (negative coefficients subtract from the total reward here):
\begin{align*}
r_{v} = \sum_{i=1}^3 v_i^2, \quad r_{w} &= \sum_{i=1}^3 w_i^2, \quad r_{a} = \sum_{i=1}^6 a_i^2
\end{align*}
For the field comparison, we set $\lambda_p = 5.0$, $\lambda_q = 2.0$, $\lambda_{stable} = 1.0$, $\lambda_v = -0.10$, $\lambda_w = -0.10$, $\lambda_a = -0.08$. The results from these experiments are zero-shot: the results are obtained without any feedback from the environment.
\section{EXPERIMENTAL RESULTS}
\subsection{Field Results}
\label{sec:fieldResults}
In a relatively calm area of Lameshur Bay, USVI, the vehicle completed the U-shaped pattern with three $4$ $m$ segments, receiving the next waypoint immediately after completing its current waypoint. Using the first reward shaping method, we test each policy (each trained with either the SDM, equivalent inertia box, or system ID drag models) $5$ times on the U-shaped pattern. The vehicle would start and end above an AprilTag placed on the seafloor, which marks a corner of the square pattern. The vehicle would then swim to the starting corner, receiving each of the next three corners in sequence as soon as $||p|| \le 0.5$ $m$. We use an EKF to fuse IMU and DVL measurements for state estimation, additionally fusing AprilTag loop closure measurements when the tags are in view. We report the mean tracking error, traversal time, and control effort on each segment. Given the goal point $\vec{x}_{g}$, the vehicle's location $\vec{x}(t)$, and the unit normal vector to the path $\vec{n}$, the instantaneous cross-track error at time $t$ and the mean cross-track error are:
\begin{align*}
e(t) &= ||\vec{n}^T(\vec{x}(k) - \vec{x}_g)||_2 \\
\bar{e}(T) &= \frac{1}{T}\sum_{k=1}^T||\vec{n}^T(\vec{x}(k) - \vec{x}_g)||_2
\end{align*}
The traversal time records the time taken to complete a segment in the mission. The control effort is calculated by first summing the squares of the PWM commands together at each time step on all six thrusters. Then zero order hold integration gives the control effort over time. Figure~\ref{fig:fieldUResults} shows median gains of $19\%$ in mean tracking error and $31\%$ in control effort compared to the equivalent inertia box model. During testing, we observed a small current moving nearly parallel to two edges of the pattern, introducing a bias into the measurements.
Finally, we show the vehicle performing the pattern at Yawzi Reef in St. John, USVI in our video submission.
\begin{figure}[t]
    \centering
    \includegraphics[width=1.0\linewidth]{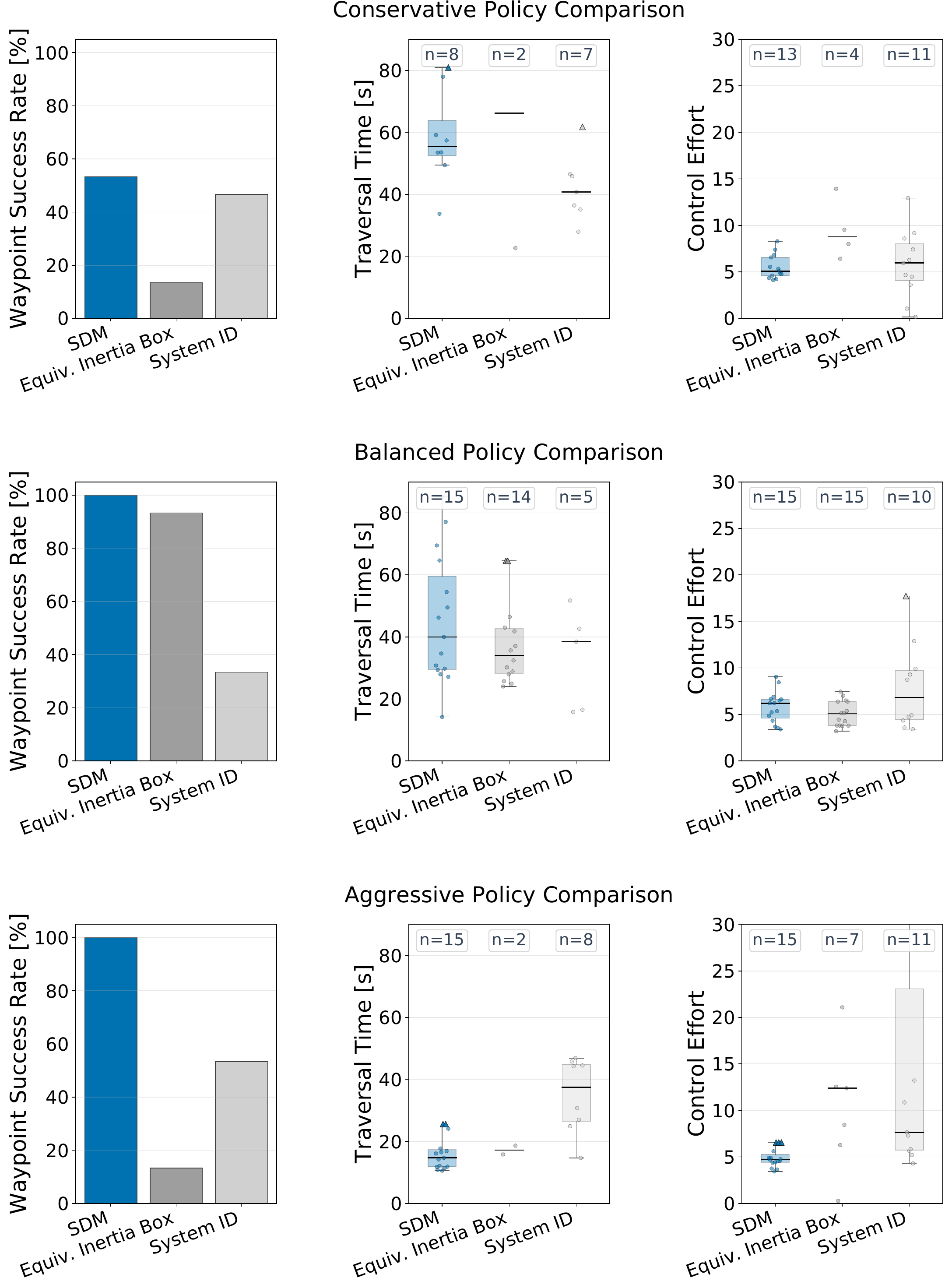}
    \caption{The SDM-based PPO policy combined with domain randomization completes the most waypoints across all three coefficient sets.}
    \label{fig:tankRewardShapingResults}
\end{figure}

\subsection{Tank Results}
\label{sec:tankResults}
In the tank experiments, we consider that a waypoint is completed when the vehicle simultaneously satisfies $||p|| \le 0.20$ $m$ and $||q|| \le 25^{\circ}$. As described in Section~\ref{sec:generalTask}, we test the following three sets of reward coefficients:
\begin{align*}
&\mathcal{C}_{conserv.} = \{\lambda_p = 0.20, \lambda_q = 0.50, \lambda_v = 0.05, \lambda_a = 0.20\} \\
&\mathcal{C}_{balanced} = \{\lambda_p = 0.40, \lambda_q = 0.90, \lambda_v = 0.00, \lambda_a = 0.20\} \\
&\mathcal{C}_{aggress.} = \{\lambda_p = 0.20, \lambda_q = 1.0, \lambda_v = 0.00, \lambda_a = 0.08\} 
\end{align*}
The first set of coefficients from the tuning experiment outlined in Section~\ref{sec:generalTask} produced controllers that often reached the correct positions slowly but frequently failed to rotate to the correct yaw. The second set of coefficients removed the velocity penalty and increased the pose reward relative to the action penalty, leading to promising controllers from the equivalent inertia box and SDM methods. The final set of coefficients aimed to reduce transit times by encouraging more thrust. We find that the equivalent inertia box based controller was highly sensitive to the reward shaping choices. The system ID based controller achieved mediocre performance across the three sets, completing more waypoints than the equivalent inertia box
\begin{figure}[t]
    \centering
    \includegraphics[width=1.0\linewidth]{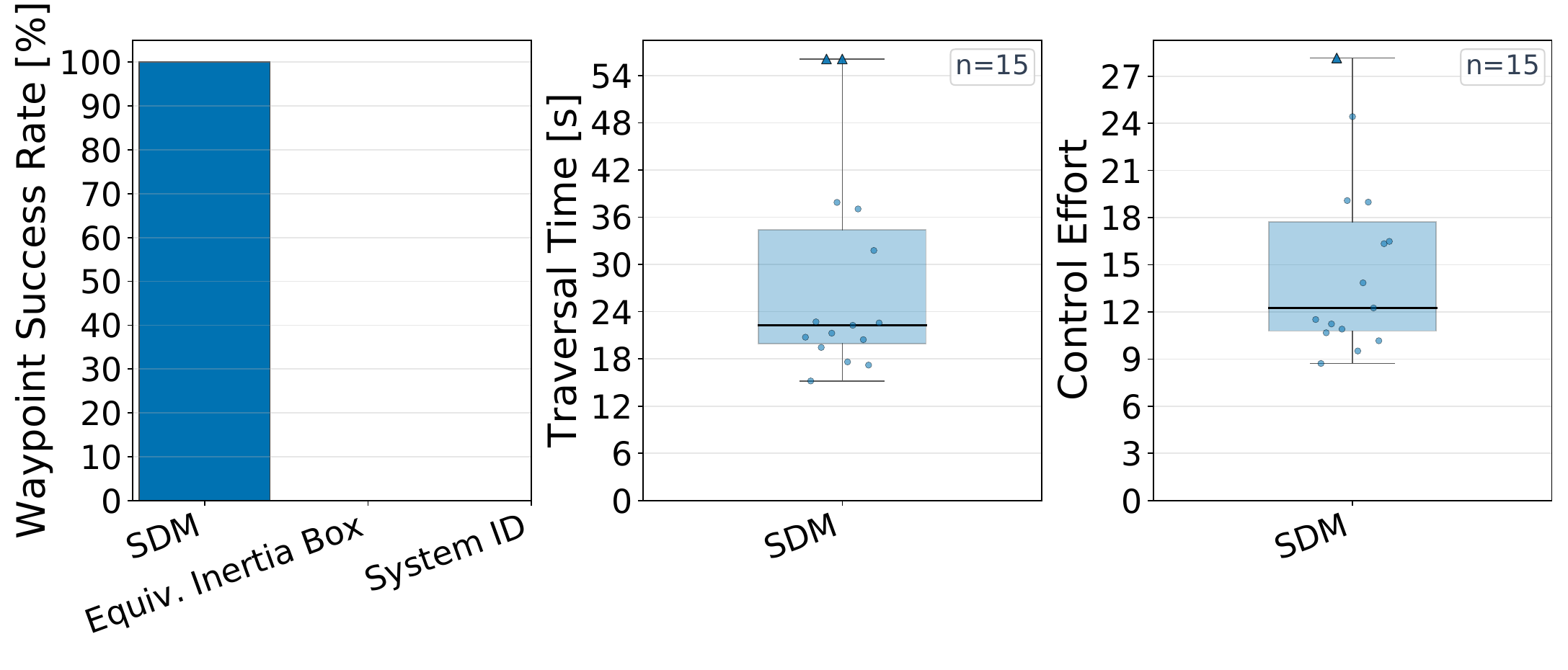}
    \caption{With an extra $0.907$ $kg$ payload on vehicle and using the associated level of domain randomization during policy training, only the SDM-based policy transfers successfully.}
    \label{fig:tankWeightResults}
\end{figure}
method with coefficient sets 1 and 3 but less with coefficient set 2. The SDM method is the most robust on the reward shaping choices: Figure~\ref{fig:tankRewardShapingResults} shows more consistent performance across all three choices. It appears that a higher-fidelity drag model leads to more consistent transfer over the three choices; the system ID model performs mediocre across all policy choices, and the SDM based model performs the best overall. We hypothesize that the cross-axis coupling in the CFD dataset (each independent case contains this) lowers the sim-to-real gap and is largely responsible for this gain. 

\begin{figure}[t]
    \centering
    \includegraphics[width=1.0\linewidth]{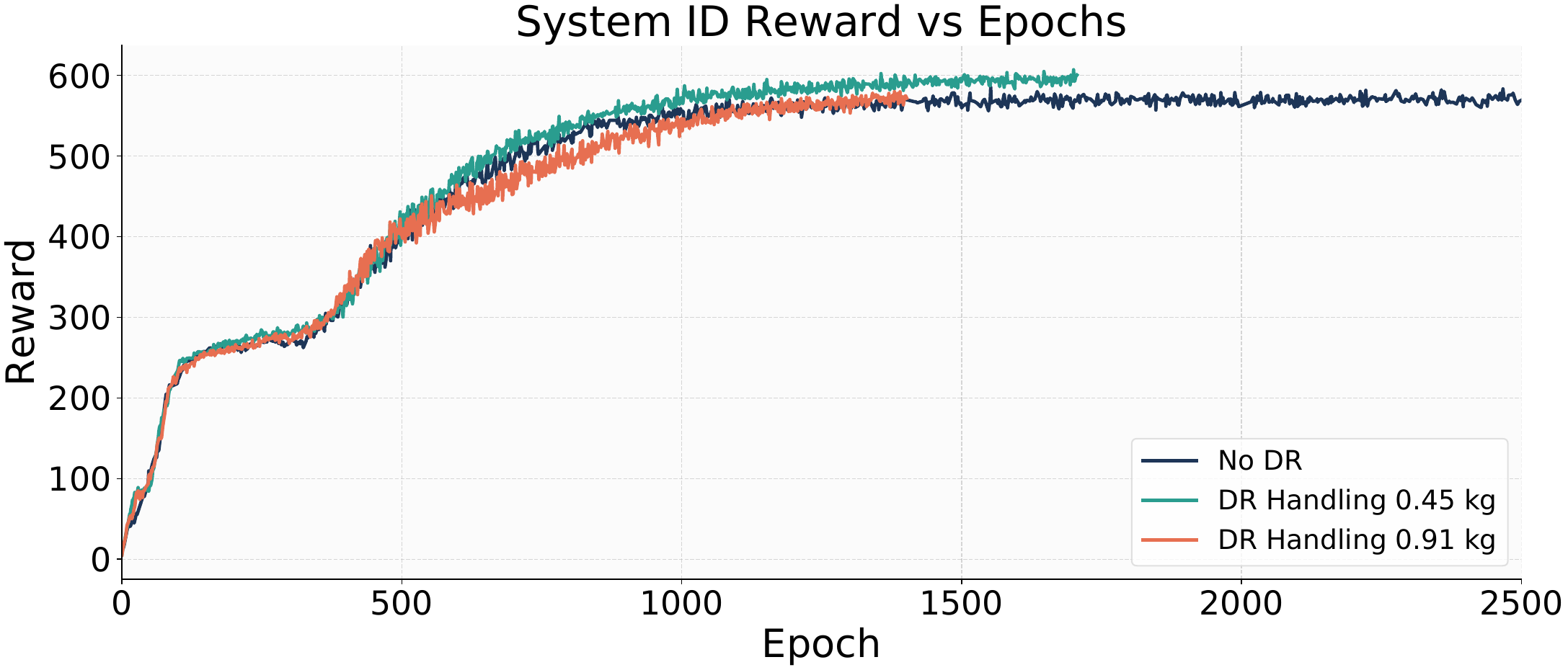}
    \caption{Increasing domain randomization on the best system ID based policy discovered in Section~\ref{sec:tankResults} leads to similar levels of reward achieved in simulation.}
    \label{fig:simSystemIDRewardCurve}
\end{figure}

\begin{figure}[t]
    \centering
    \includegraphics[width=1.0\linewidth]{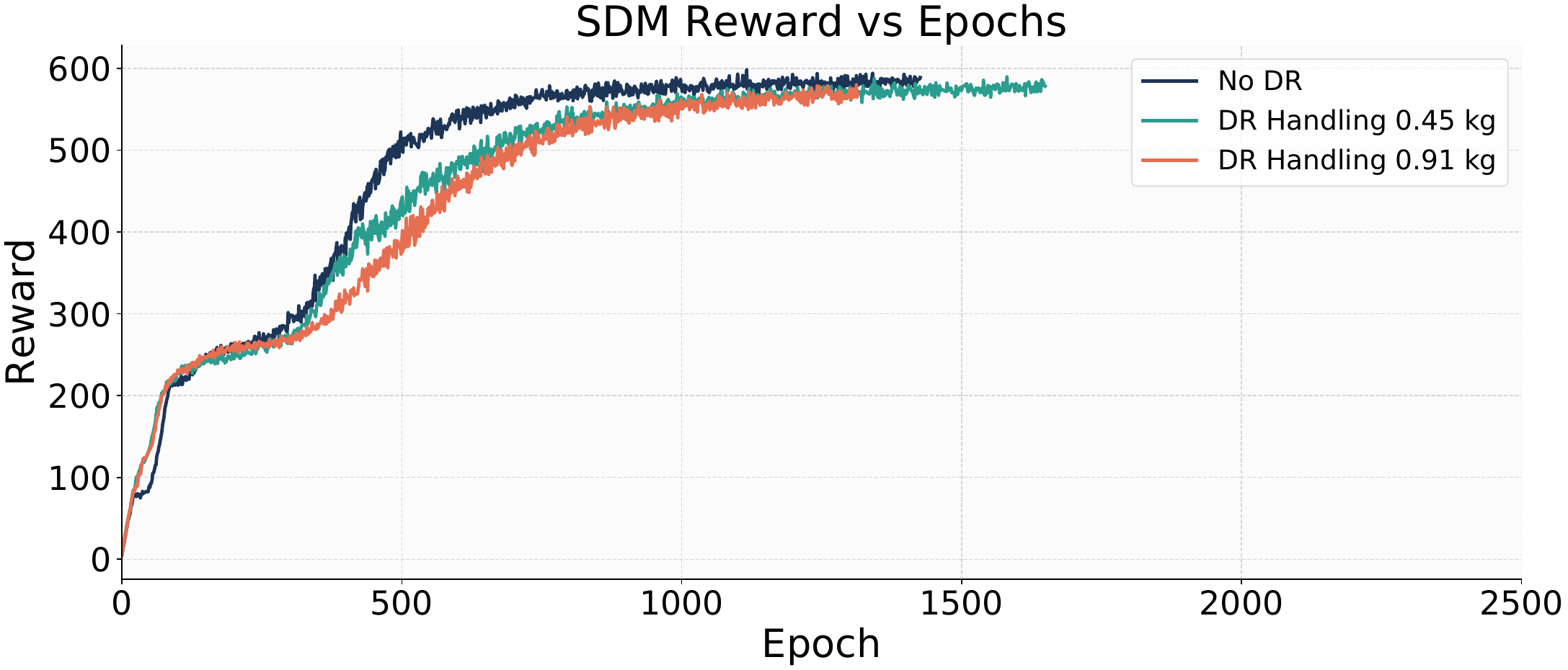}
    \caption{Increasing domain randomization on the best SDM-based policy discovered in Section~\ref{sec:tankResults} leads to similar levels of reward achieved in simulation.}
    \label{fig:simCFDRewardCurve}
%
    \centering
    \includegraphics[width=1.0\linewidth]{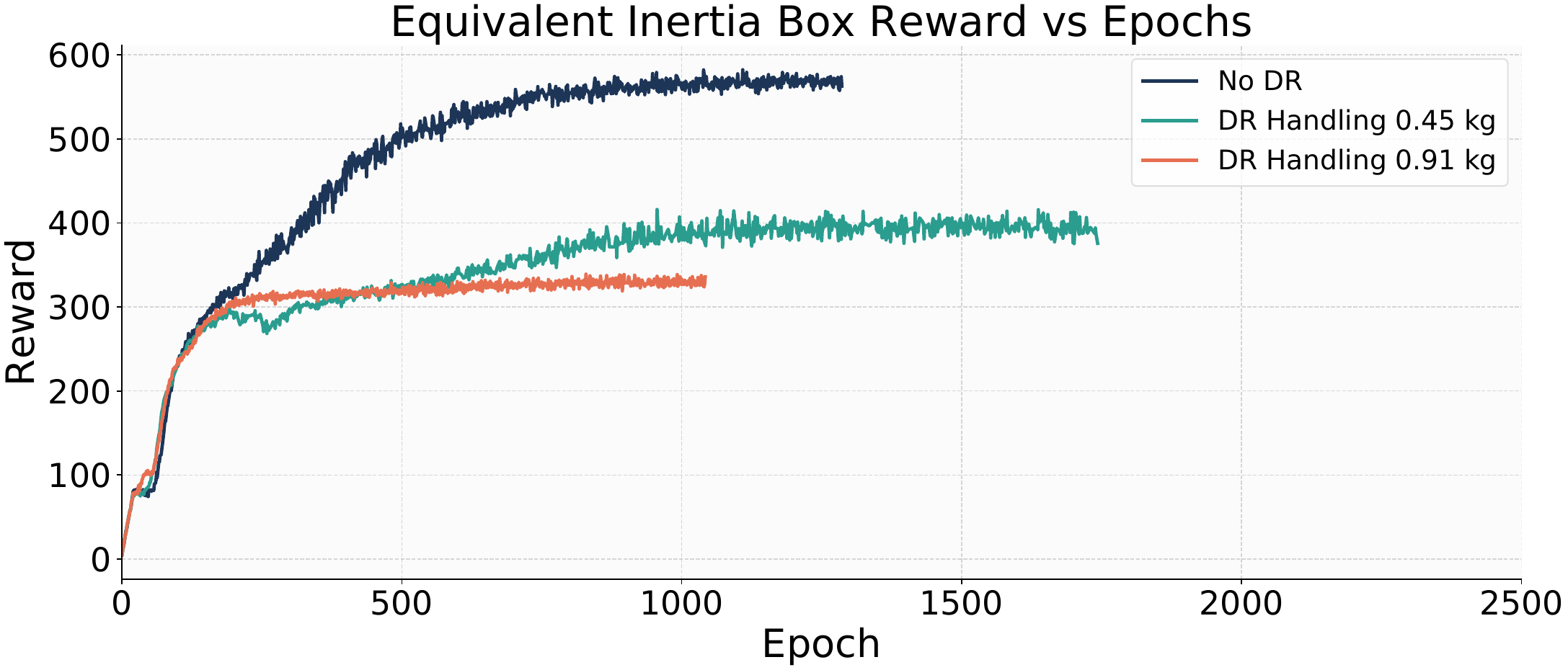}
    \caption{Increasing domain randomization on the best equivalent inertia box based policy discovered in Section~\ref{sec:tankResults} leads to significant loss of total reward achieved.}
    \label{fig:simBoxRewardCurve}
\end{figure}

Comparing the best policies discovered with each drag model assumption, we find that the SDM-based policy completed the most waypoints, had the fastest traversal times, and had the lowest control effort (Figure~\ref{fig:tankBestResults}). This is consistent with our findings in the field experiments (Figure~\ref{fig:fieldUResults}).
Taking the simulation configuration that produced the best policies displayed in Figure~\ref{fig:tankBestResults}, we desired to see if domain randomization could enable the policies to handle an additional $0.907$ $kg$ of mass. We calculated that $\pm2\%$ in volume DR and $0.02$ $m$ variation in the COB-COM offset is sufficient to handle an additional $0.907$ $kg$ mass on the vehicle. We thus added that level of DR to the training process. 
As seen in Figures~\ref{fig:simSystemIDRewardCurve} and \ref{fig:simCFDRewardCurve}, the SDM and system ID simulations achieved rewards similar to those obtained without DR, whereas the equivalent inertia box model experienced a loss of nearly half the total reward. These results suggested that the SDM and system ID policies might transfer successfully to the vehicle, while the equivalent inertia box policy would not. Additionally, it seems that learning is highly sensitive to drag coefficients. The equivalent inertia box and system ID methods have the same diagonal drag structure, yet they achieve varying reward as DR is introduced. Future work may investigate this relationship more fully.

When the three policies were deployed on the vehicle with an additional $0.907$ $kg$ of mass, Figure~\ref{fig:tankWeightResults} shows that the SDM-based policy completed the U-shaped task in all five trials, although it was slower and required more thrust than without the additional mass. In contrast, neither of the other two policies completed any of the waypoints. Thus, the reward curve in Figure~\ref{fig:simCFDRewardCurve} correctly predicted successful transfer of the SDM-based policy, whereas the high simulated reward of the system ID policy did not translate to successful deployment. This provides evidence that the use of SDMs in the RL training pipeline leads to more accurate predictions of zero-shot transfer.

\section{CONCLUSION}
\label{sec:conclusion}
The experiments demonstrate that dynamics-model fidelity can remain critical to sim-to-real RL even when domain randomization is used. Across field tracking, reward-design variations, and a substantial physical perturbation, policies trained with the CFD-derived surrogate transferred more consistently than policies trained with lower-fidelity models. Most notably, under the $0.907$ $kg$ perturbation, simulation predicted successful transfer for both the system-ID and SDM models, but only the SDM-based policy transferred successfully on hardware. These results demonstrate that domain randomization may not compensate for important inaccuracies in the nominal dynamics model, and that improving model fidelity can improve both policy performance and the predictive value of simulation.



\section*{ACKNOWLEDGMENT}
This work was supported in part by generous WHOI donors, and Steven Roche was supported by the DoD SMART Scholarship. Special thanks to Eric Chen, Daniel Yang, Sierra Jarriel, Trevor Milliken, and Gemma Luther for helping with field experiments. Thank you to John Walsh and John Cast for help with hardware troubleshooting and to Robert McCabe for scheduling tank testing. Codex was used to produce data visualization in figures~\ref{fig:fieldUResults}--\ref{fig:simBoxRewardCurve}, for boilerplate for the SDMs, and for boilerplate in code that solves the CFD simulations.

\bibliographystyle{IEEEtran}
\bibliography{IEEEabrv,references}

@article{c1,
  author    = {J. Das and F. Py and J. B. Harvey and J. P. Ryan and A. Gellene and R. Graham and D. A. Caron and K. Rajan and G. S. Sukhatme},
  title     = {Data-driven robotic sampling for marine ecosystem monitoring},
  journal   = {International Journal of Robotics Research},
  year      = {2015}
}

@article{c2,
  author    = {L. Cai and N. E. McGuire and R. Hanlon and T. A. Mooney and Y. Girdhar},
  title     = {Semi-supervised Visual Tracking of Marine Animals Using Autonomous Underwater Vehicles},
  journal   = {International Journal of Computer Vision},
  year      = {2023}
}

@article{c3,
  author    = {D. R. Yoerger and A. F. Govindarajan and J. C. Howland and J. K. Llopiz and P. H. Wiebe and M. Curran and J. Fujii and D. Gomez-Ibanez and K. Katija and B. H. Robison and B. W. Hobson and M. Risi and S. M. Rock},
  title     = {A hybrid underwater robot for multidisciplinary investigation of the ocean twilight zone},
  journal   = {Science Robotics},
  year      = {2021}
}

@inproceedings{c4,
  author    = {T. Manderson and J. C. G. Higuera and S. Wapnick and J. F. Tremblay and F. Shkurti and D. Meger and G. Dudek},
  title     = {Vision-Based Goal-Conditioned Policies for Underwater Navigation in the Presence of Obstacles},
  booktitle = {Robotics: Science and Systems},
  year      = {2020}
}

@inproceedings{c5,
  author    = {R. Vivekanandan and D. Chang and G. A. Hollinger},
  title     = {Autonomous Underwater Docking using Flow State Estimation and Model Predictive Control},
  booktitle = {Proceedings of the IEEE International Conference on Robotics and Automation (ICRA)},
  year      = {2023}
}

@inproceedings{c6,
  author    = {M. Xanthidis and others},
  title     = {Navigation in the Presence of Obstacles for an Agile Autonomous Underwater Vehicle},
  booktitle = {Proceedings of the IEEE International Conference on Robotics and Automation (ICRA)},
  year      = {2020}
}

@inproceedings{c7,
  author    = {Y. Girdhar and N. McGuire and L. Cai and S. Jamieson and S. McCammon and B. Claus and J. E. San Soucie and J. E. Todd and T. A. Mooney},
  title     = {CUREE: A Curious Underwater Robot for Ecosystem Exploration},
  booktitle = {Proceedings of the IEEE International Conference on Robotics and Automation (ICRA)},
  year      = {2023}
}

@article{c8,
  author    = {C. Li and S. Guo and J. Guo},
  title     = {Study on Obstacle Avoidance Strategy Using Multiple Ultrasonic Sensors for Spherical Underwater Robots},
  journal   = {IEEE Sensors Journal},
  volume    = {22},
  number    = {24},
  year      = {2022}
}

@article{c9,
  author    = {S. Bhat and C. Panteli and I. Stenius and D. V. Dimarogonas},
  title     = {Nonlinear model predictive control for hydrobatics: Experiments with an underactuated AUV},
  journal   = {Journal of Field Robotics},
  year      = {2023}
}

@inproceedings{c10,
  author    = {A. Mitchell and E. McGookin and D. Murray-Smith},
  title     = {Comparison of control methods for autonomous underwater vehicles},
  booktitle = {IFAC Workshop on Guidance and Control of Underwater Vehicles},
  journal   = {IFAC Proceedings Volumes},
  volume    = {36},
  number    = {4},
  pages     = {37--42},
  year      = {2003},
  address   = {Newport, South Wales, UK},
  url       = {https://www.sciencedirect.com/science/article/pii/S1474667017366545}
}

@misc{molchanov2019simtomultirealtransferlowlevelrobust,
      title={Sim-to-(Multi)-Real: Transfer of Low-Level Robust Control Policies to Multiple Quadrotors}, 
      author={Artem Molchanov and Tao Chen and Wolfgang Hönig and James A. Preiss and Nora Ayanian and Gaurav S. Sukhatme},
      year={2019},
      eprint={1903.04628},
      archivePrefix={arXiv},
      primaryClass={cs.RO},
      url={https://arxiv.org/abs/1903.04628}, 
}

@inproceedings{c11_anonymized,
  author    = {Anonymized},
  title     = {Anonymized},
  booktitle = {Proceedings of the IEEE International Conference on Robotics and Automation (ICRA)},
  year      = {2025}
}

@article{tensorialApproachCompCont,
  author    = {H. G. Weller and G. Tabor and H. Jasak and C. Fureby},
  title     = {A tensorial approach to computational continuum mechanics using object-oriented techniques},
  journal   = {Computers in Physics},
  volume    = {12},
  number    = {6},
  year      = {1998}
}

@article{weng_autonomous_2024,
    title = {Autonomous underwater vehicle link alignment control in unknown environments using reinforcement learning},
    volume = {41},
    copyright = {© 2024 The Authors. Journal of Field Robotics published by Wiley Periodicals LLC.},
    issn = {1556-4967},
    url = {https://onlinelibrary.wiley.com/doi/abs/10.1002/rob.22348},
    language = {en},
    number = {6},
    urldate = {2025-09-12},
    journal = {Journal of Field Robotics},
    author = {Weng, Yang and Chun, Sehwa and Ohashi, Masaki and Matsuda, Takumi and Sekimori, Yuki and Pajarinen, Joni and Peters, Jan and Maki, Toshihiro},
    year = {2024},
    pages = {1724--1743},
}

@article{liu_deep_2021,
    title = {Deep {Reinforcement} {Learning} for {Vectored} {Thruster} {Autonomous} {Underwater} {Vehicle} {Control}},
    volume = {2021},
    copyright = {Copyright © 2021 Tao Liu et al.},
    issn = {1099-0526},
    url = {https://onlinelibrary.wiley.com/doi/abs/10.1155/2021/6649625},
    doi = {10.1155/2021/6649625},
    language = {en},
    number = {1},
    urldate = {2025-09-12},
    journal = {Complexity},
    author = {Liu, Tao and Hu, Yuli and Xu, Hui},
    year = {2021},
    pages = {6649625},
}

@misc{noauthor_openfoam_2025,
    title = {{OpenFOAM}},
    url = {https://www.openfoam.com/},
    language = {en},
    urldate = {2025-09-12},
    month = jun,
    year = {2025},
}

@misc{schulman_proximal_2017,
    title = {Proximal {Policy} {Optimization} {Algorithms}},
    url = {http://arxiv.org/abs/1707.06347},
    doi = {10.48550/arXiv.1707.06347},
    urldate = {2025-05-01},
    publisher = {arXiv},
    author = {Schulman, John and Wolski, Filip and Dhariwal, Prafulla and Radford, Alec and Klimov, Oleg},
    month = aug,
    year = {2017},
    note = {arXiv:1707.06347 [cs]},
}

@misc{tunçay2026fastpolicylearning6dof,
      title={Fast Policy Learning for 6-DOF Position Control of Underwater Vehicles}, 
      author={Sümer Tunçay and Alain Andres and Ignacio Carlucho},
      year={2026},
      eprint={2512.13359},
      archivePrefix={arXiv},
      primaryClass={cs.RO},
      url={https://arxiv.org/abs/2512.13359}, 
}

@misc{fosso2025sim2swimzeroshotvelocitycontrol,
      title={Sim2Swim: Zero-Shot Velocity Control for Agile AUV Maneuvering in 3 Minutes}, 
      author={Lauritz Rismark Fosso and Herman Biørn Amundsen and Marios Xanthidis and Sveinung Johan Ohrem},
      year={2025},
      eprint={2512.08656},
      archivePrefix={arXiv},
      primaryClass={cs.RO},
      url={https://arxiv.org/abs/2512.08656}, 
}

@article{pointToPointNavigationFishSwimmer,
	author = {Zhu, Yi and Pang, Jian-Hua and Tian, Fang-Bao},
	doi = {10.3389/fphy.2022.870273},
	issn = {2296-424X},
	journal = {Frontiers in Physics},
	title = {Point-to-Point Navigation of a Fish-Like Swimmer in a Vortical Flow With Deep Reinforcement Learning},
	url = {https://www.frontiersin.org/journals/physics/articles/10.3389/fphy.2022.870273},
	volume = {Volume 10 - 2022},
	year = {2022}}

@article{RLforTurbulentFlows,
	author = {Artur K. Lidtke and Douwe Rijpkema and B{\"u}lent D{\"u}z},
	doi = {https://doi.org/10.1016/j.oceaneng.2024.118538},
	issn = {0029-8018},
	journal = {Ocean Engineering},
	pages = {118538},
	title = {General reinforcement learning control for AUV manoeuvring in turbulent flows},
	url = {https://www.sciencedirect.com/science/article/pii/S0029801824018766},
	volume = {309},
	year = {2024}}

@article{Cui_2024,
   title={Enhancing efficiency and propulsion in bio-mimetic robotic fish through end-to-end deep reinforcement learning},
   volume={36},
   ISSN={1089-7666},
   url={http://dx.doi.org/10.1063/5.0192993},
   DOI={10.1063/5.0192993},
   number={3},
   journal={Physics of Fluids},
   publisher={AIP Publishing},
   author={Cui, Xinyu and Sun, Boai and Zhu, Yi and Yang, Ning and Zhang, Haifeng and Cui, Weicheng and Fan, Dixia and Wang, Jun},
   year={2024},
   month=Mar }

@article{learningFrameworkForFishRobots,
author = {Zhang, Tianhao and Tian, Runyu and Yang, Hongqi and Wang, Chen and Sun, Jinan and Zhang, Shikun and Xie, Guangming},
year = {2022},
month = {12},
pages = {1-18},
title = {From Simulation to Reality: A Learning Framework for Fish-Like Robots to Perform Control Tasks},
volume = {PP},
journal = {IEEE Transactions on Robotics},
doi = {10.1109/TRO.2022.3181014}
}

@misc{lin2025learningagileswimmingendtoend,
      title={Learning Agile Swimming: An End-to-End Approach without CPGs}, 
      author={Xiaozhu Lin and Xiaopei Liu and Yang Wang},
      year={2025},
      eprint={2409.10019},
      archivePrefix={arXiv},
      primaryClass={cs.RO},
      doi={https://doi.org/10.1109/LRA.2025.3527757},
      url={https://arxiv.org/abs/2409.10019}, 
}

@article{Todorov2012MuJoCoAP,
  title={MuJoCo: A physics engine for model-based control},
  author={Emanuel Todorov and Tom Erez and Yuval Tassa},
  journal={2012 IEEE/RSJ International Conference on Intelligent Robots and Systems},
  year={2012},
  pages={5026-5033},
  url={https://api.semanticscholar.org/CorpusID:5230692}
}

@misc{liu2022fishgymhighperformancephysicsbasedsimulation,
      title={FishGym: A High-Performance Physics-based Simulation Framework for Underwater Robot Learning}, 
      author={Wenji Liu and Kai Bai and Xuming He and Shuran Song and Changxi Zheng and Xiaopei Liu},
      year={2022},
      eprint={2206.01683},
      archivePrefix={arXiv},
      primaryClass={cs.RO},
      url={https://arxiv.org/abs/2206.01683}, 
}

@article{swim4Real,
  author={Sufán, Vicente and Troni, Giancarlo},
  journal={IEEE Robotics and Automation Letters}, 
  title={Swim4Real: Deep Reinforcement Learning-Based Energy-Efficient and Agile 6-DOF Control for Underwater Vehicles}, 
  year={2025},
  volume={10},
  number={7},
  pages={7326-7333},
  doi={10.1109/LRA.2025.3575650}}

@article{SUN2026125161,
title = {Deep reinforcement learning control of fish locomotion via physics-guided surrogates},
journal = {Ocean Engineering},
volume = {355},
pages = {125161},
year = {2026},
issn = {0029-8018},
doi = {https://doi.org/10.1016/j.oceaneng.2026.125161},
url = {https://www.sciencedirect.com/science/article/pii/S0029801826009959},
author = {Weiyuan Sun and Ruixin Zhan and Yunfei Wang and Haokui Jiang and Juntian Qu and Xiaofan Li and Shunxiang Cao}
}

@INPROCEEDINGS{11509187,
  author={Xu, Zikang and Zhou, Lei and Zheng, Yawei and Zhu, An and Ai, Haiping},
  booktitle={2026 9th International Conference on Advanced Algorithms and Control Engineering (ICAACE)}, 
  title={Research on Hydrodynamic Modeling and Control of Underwater Robots Based on Deep Learning}, 
  year={2026},
  volume={},
  number={},
  pages={2163-2168},
  doi={10.1109/ICAACE69793.2026.11509187}}

@article{reinforcementLearningSurvey2026,
title = {Reinforcement learning in robotic systems : A review on sim-to-real transfer},
journal = {Robotics and Autonomous Systems},
volume = {198},
pages = {105327},
year = {2026},
issn = {0921-8890},
doi = {https://doi.org/10.1016/j.robot.2025.105327},
url = {https://www.sciencedirect.com/science/article/pii/S0921889025004245},
author = {Rajesh Tiwari and Shailesh Khapre and Avantika Singh}
}

@InProceedings{pmlr-v164-rudin22a,
  title = 	 {Learning to Walk in Minutes Using Massively Parallel Deep Reinforcement Learning},
  author =       {Rudin, Nikita and Hoeller, David and Reist, Philipp and Hutter, Marco},
  booktitle = 	 {Proceedings of the 5th Conference on Robot Learning},
  pages = 	 {91--100},
  year = 	 {2022},
  editor = 	 {Faust, Aleksandra and Hsu, David and Neumann, Gerhard},
  volume = 	 {164},
  series = 	 {Proceedings of Machine Learning Research},
  month = 	 {08--11 Nov},
  publisher =    {PMLR},
  url = 	 {https://proceedings.mlr.press/v164/rudin22a.html}
}

@software{NVIDIA_Isaac_Sim,
author = {{NVIDIA}},
license = {Apache-2.0},
title = {{Isaac Sim}},
url = {https://github.com/isaac-sim/IsaacSim},
version = {6.0.1}
}

\addtolength{\textheight}{-12cm}   

\end{document}